\documentclass[10pt,twocolumn,letterpaper]{article}

\usepackage{cvpr}
\usepackage{times}
\usepackage{epsfig}
\usepackage{graphicx}
\usepackage{amsmath}
\usepackage{amssymb}

\usepackage{animate}
\usepackage{multirow}

\usepackage[pagebackref=true,breaklinks=true,letterpaper=true,colorlinks,bookmarks=false]{hyperref}

\cvprfinalcopy 

\def\cvprPaperID{23} 

\ifcvprfinal\fi
\begin{document}

\title{VPUTeam: A Prospective Study on Sequence-Driven Temporal Sampling and \\ Ego-Motion Compensation for Action Recognition in the EPIC-Kitchens Dataset}

\author{Alejandro L\'{o}pez-Cifuentes\\
Video Processing and Understanding Lab\\
Universidad Aut\'{o}noma de Madrid\\
{\tt\small alejandro.lopezc@uam.es}
\and
Marcos Escudero-Vi\~nolo\\
Video Processing and Understanding Lab\\
Universidad Aut\'{o}noma de Madrid\\
{\tt\small marcos.escudero@uam.es}
\and
Jes\'{u}s~Besc\'{o}s\\
Video Processing and Understanding Lab\\
Universidad Aut\'{o}noma de Madrid\\
{\tt\small j.bescos@uam.es}
}

\maketitle

\begin{abstract}
   Action recognition is currently one of the top-challenging research fields in computer vision. Convolutional Neural Networks (CNNs) have significantly boosted its performance but rely on fixed-size spatio-temporal windows of analysis, reducing CNNs temporal receptive fields. Among action recognition datasets, egocentric recorded sequences have become of important relevance while entailing an additional challenge: ego-motion is unavoidably transferred to these sequences. The proposed method aims to cope with it by estimating this ego-motion or camera motion. The estimation is used to temporally partition video sequences into motion-compensated temporal \textit{chunks} showing the action under stable backgrounds and allowing for a content-driven temporal sampling. A CNN trained in an end-to-end fashion is used to extract temporal features from each \textit{chunk}, which are late fused. This process leads to the extraction of features from the whole temporal range of an action, increasing the temporal receptive field of the network.
   \textcolor{blue}{This document is best viewed offline where some figures play as animation.\footnote{You can find a dynamic preprint version of the paper at: \href{https://github.com/vpulab/EPIC-Kitchens-2020-Subset}{Dynamic Version}}}
\end{abstract}

\section{Introduction}
Video action classification is a highly emerging research topic in computer vision \cite{kay2017kinetics, carreira2018short, sigurdsson2016hollywood} due to its potential wide range of applications.

Among reported approaches, those relying on the use of Convolutional Neural Networks (CNNs) have reported the highest performances. These methods are based on the extraction of spatio-temporal features on the video, which are then used to classify the action recorded. Due to processing constraints, these features are instead usually extracted on video segments—i.e. fixed-size spatio-temporal windows obtained by sampling and cutting the video, hence discarding the rest of the video sequence. CNNs with reduced temporal receptive fields emerge from this design criteria.

\begin{figure}[t!]
    \centering
    \includegraphics[width=\linewidth,keepaspectratio]{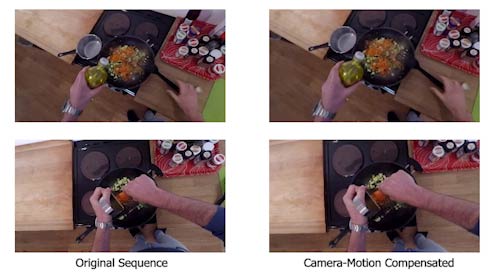}
    \caption{Left Column: Original video sequences corresponding to actions \textit{Pour Oil} and \textit{Grating Carrot}. Right Column: Results of the proposed camera motion compensation approach with unsupervised clustering. \textcolor{blue}{Best viewed with zoom in Adobe Reader where figure should play as videos.}}
    \label{fig:Compensated Video}
\end{figure}

\begin{figure*}[ht!]
    \centering
    \includegraphics[width=\linewidth,keepaspectratio]{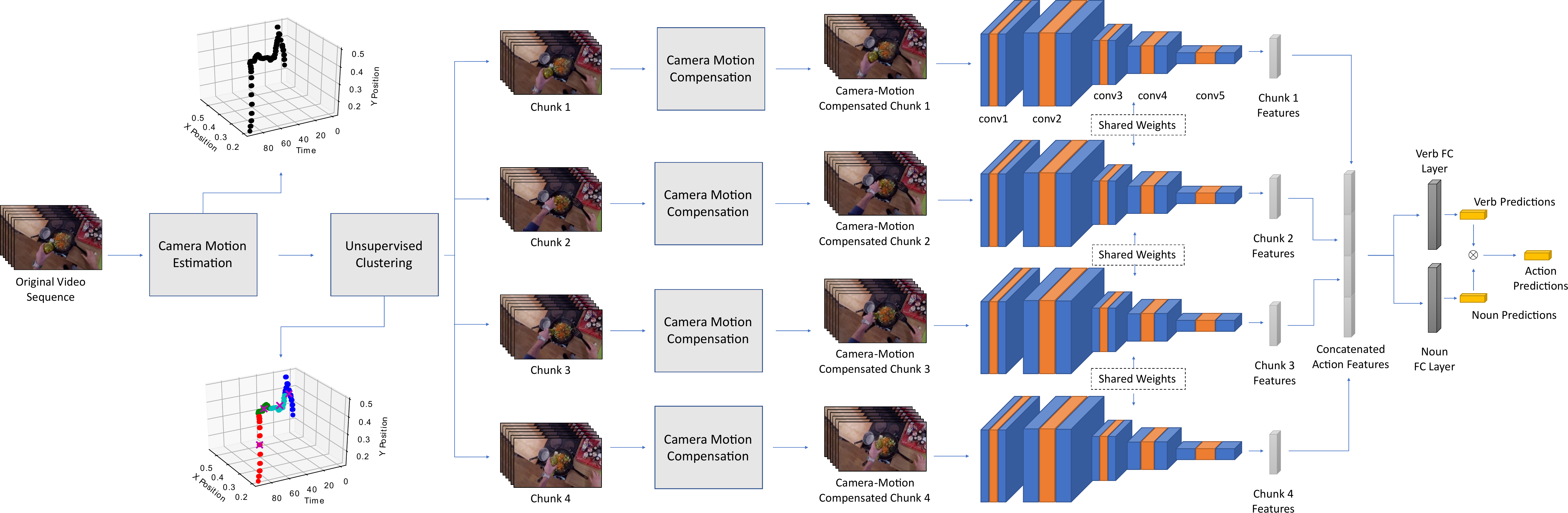}
    \caption{Proposed architecture for the action classification task. Camera motion is estimated for a given action sequence. Using this estimation an unsupervised 3D clustering based on spatial and temporal positions of the camera is performed, dividing the action sequence into temporal \textit{chunks}. Camera motion is compensated independently for each chunk leading to quasi-static sub-sequences. Resulting \textit{chunks} are fed to a 3D CNN which obtains features for each of them. Late fusion aims to gather features from all the \textit{chunks} to obtain the final predictions for verb, noun and action.}
    \label{fig:Proposed Architecture}
\end{figure*}

There are several video action classification datasets in the literature, among them, egocentric ones, i.e. those recorded from a first person point of view, have become of crucial importance \cite{damen2018scaling}. The egocentric domain presents a set of relevant advantages with respect to third-person recordings. The point of view in egocentric videos, closely related to the position of humans eyes, provides a view similar to what of what we humans actually see. Besides, the close proximity of the wearable camera to the undergoing action provides closer and more detailed objects representations compared to third-person view recordings.

However, egocentric recordings entail an additional challenge: human body or ego-motion is inevitably transferred to video sequences (see left column in Figure \ref{fig:Compensated Video}), creating motion and temporal patterns that may occlude or befoul the action's ones. This ego-motion might hinder the performance of modern CNNs trained to recognize actions in videos \cite{price2019evaluation,feichtenhofer2019slowfast, wang2018non,wang2019baidu} as these are focused on extracting the representative temporal features defining an action.

Our proposal aims to cope with ego-motion problems while providing a sequence-driven adaptive temporal sampling scheme. This process aims to filter out the majority of the ego-motion, leading to an action segment with a quasi-static background. Camera motion compensation accentuates the representative motion in a specific action, such as the movement of objects or hands (Figure \ref{fig:Compensated Video}), which might lead to more representative and distinguishable action features. In addition, we use the camera motion estimation to temporally divide the sequence into representative context temporal \textit{chunks}. This temporal partitioning benefits the feature extraction process by: 1) Easing the process of camera motion compensation by having \textit{chunks} that represent contextually similar backgrounds with limited camera motion. 2) Enabling the use of an irregular temporal sampling policy hence expanding the CNN temporal receptive field. We adopt a shared end-to-end fashion CNN to independently analyze each of the camera motion compensated \textit{chunks}. Finally, features from each chunk are late fused to obtain final action predictions.

\begin{figure}[t!]
    \centering
    \includegraphics[width=0.8\linewidth,keepaspectratio]{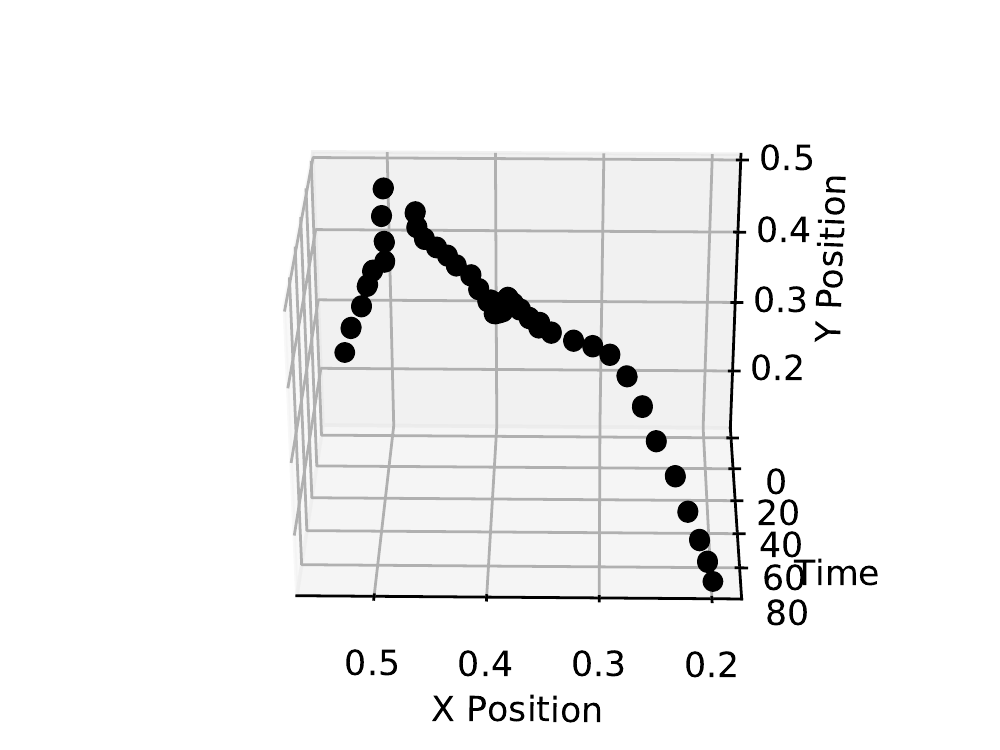}
    \caption{Camera motion estimation example. Each black dot corresponds to the projected middle coordinates of a frame using and homography with respect to a reference frame. \(X\) and \(Y\) axes represent normalized spatial movement while \(Z\) axis represents time in frame scale.  \textcolor{blue}{Best viewed in Adobe Reader where figure should play as animation.}}
    \label{fig:camera motion Estimation}
\end{figure}

\section{Method}
Four different stages are posed for the action classification task. First, camera motion is estimated for an action video sequence (Section \ref{subsec:Camera Motion Estimation}). Then, the resulting estimation is used to partition the original video sequence into temporal \textit{chunks} via unsupervised clustering (Section \ref{subsec:Unsupervised Chunk Division}). Camera motion is then compensated for each temporal chunk independently (Section \ref{subsec:Camera Motion Compensation}). Finally, motion compensated \textit{chunks} are fed to a CNN which obtains the final predictions (Section \ref{subsec:CNN Architecture}). The proposed pipeline is depicted and detailed in Figure \ref{fig:Proposed Architecture}.

\subsection{Camera Motion Estimation} \label{subsec:Camera Motion Estimation}
We estimate the camera motion of a given sequence by using a classical stereo image rectification technique \cite{hartley2003multiple}. First, a pre-trained D2Net CNN \cite{dusmanu2019d2} is used to obtain reliable pixel-level features for each video frame. Features of every frame are then matched to the ones from a chosen reference one\textemdash typically the first or the last frame. These matching correspondences are used to compute planar homographies using RANSAC \cite{fischler1981random}. So-extracted homographies define the transforms between every frame and the reference frame. An estimation of the camera motion during the sequence is obtained by projecting the middle point coordinates from every frame using the corresponding homography (Figure \ref{fig:camera motion Estimation}).

Video sequences might include high camera motion in terms of panning or tilting. This severe motion leads to high background variation along frames, hindering computation of a motion-compensated sequence. To overcome this issue we propose to divide each video sequences into different temporal chunks using unsupervised clustering. 

\begin{figure}[t!]
    \centering
    \includegraphics[width=0.8\linewidth,keepaspectratio]{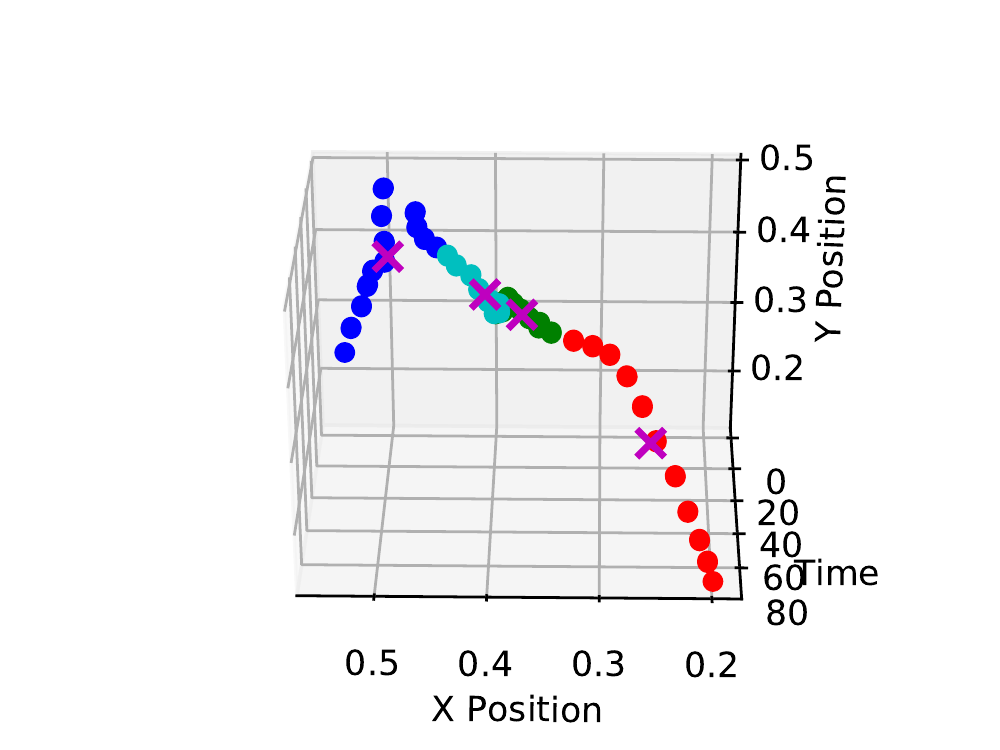}
    \caption{Unsupervised clustering of a video sequence into different \textit{chunks} relying on camera motion estimation. This figure represents an example where 4 \textit{chunks} are used. \textit{Chunks} are represented using different color dots. Purple \(X\) markers represent the cluster center. \textcolor{blue}{Best viewed with zoom in Adobe Reader where figure should play as animation.}}
    \label{fig:Unsupervised Clustering}
\end{figure}

\subsection{Unsupervised Partition in Temporal Chunks} \label{subsec:Unsupervised Chunk Division}
Unsupervised clustering is performed using the KMeans technique \cite{kanungo2002efficient} fed by the spatio-temporal coordinates extracted in the camera motion estimation stage. By this, the temporal span of a video sequence is reduced and the original video sequence is divided into smaller parts with contextually similar backgrounds. An example of this process is depicted in Figure \ref{fig:Unsupervised Clustering}.

\subsection{Camera Motion Compensation} \label{subsec:Camera Motion Compensation}
The image rectification process described in section \ref{subsec:Camera Motion Estimation} is performed individually for each chunk reusing the features there extracted but defining a new reference frame for each chunk. Using the new obtained homographies, the frames in a \textit{chunk} are warped to its \textit{chunk}-reference frame to a camera motion compensated chunk (Figure \ref{fig:Compensated Video} right column). These compensated \textit{chunks} are the ones used for the end-to-end training of the shared Convolutional Neural Network (CNN). The proposed technique for camera motion compensation and sequence partition into chunks is highly improvable. This is a first approach to prove that this benefits the action recognition process. 

\subsection{CNN Architecture} \label{subsec:CNN Architecture}
To obtain temporal features from \textit{chunks} the Slow-Fast ResNet-50 CNN \cite{feichtenhofer2019slowfast} is used as backbone. The Slow-Fast is composed by the Spatial branch and the Temporal branch. For our participation, due to some computational constraints, we have only explored the use of the Temporal branch CNN. This branch expands temporal channels while reducing the spatial ones, ensuring that final features are closer to the temporal domain than to the spatial one. 

Each chunk is independently fed-forwarded to the same 3D CNN to obtain features. Chunk features are later fused through concatenation to obtain a final feature vector. The use of \textit{chunks} leads to the extraction of features from the whole temporal range of the action, hence increasing the temporal receptive field of the network with respect to analyzing only a fixed-size temporal window.

Two different fully-connected layers and regular logarithmic softmax are used to obtain verb (\(v\)) and noun (\(n\)) probabilities \(p_v\) and \(p_n\) respectively. \(p_v\) and \(p_n\) are used to obtain action predictions as:
\begin{equation}
    p_a = p_v \otimes p_n,
\end{equation}

where \((\otimes)\) is defined as the outer product between two vectors.

\begin{table*}[!t]
    \begin{centering}
    \renewcommand{\arraystretch}{1.2}
    \footnotesize
    \caption{Preliminary results. (\%)}
    \begin{tabular}{c c c c c c c c c c c c c c }
        \hline 
        \multirow{2}{*}{Method} & \multirow{2}{*}{Number of Parameters} & \multicolumn{3}{c}{Top@1} & \multicolumn{3}{c}{Top@5}\tabularnewline
         &  & Action & Verb & Noun & Action & Verb & Noun\tabularnewline
        \hline
        Baseline & 0.8 M & 29.00 \% & 52.44 \% & 40.78 \% & 48.22 \% & \textbf{85.56 \%} & \textbf{67.78 \%}\tabularnewline
        Ours & 1 M & \textbf{30.25 \%} & \textbf{54.56 \%} & \textbf{42.67 \%} & \textbf{49.67 \%} & 85.33 \% & 66.89 \%\tabularnewline
        \hline 
    \end{tabular}
    \label{tab:Preliminary Results}
    \par
    \end{centering}
\end{table*}

\begin{table*}[!t]
    \begin{centering}
    \renewcommand{\arraystretch}{1.2}
    \footnotesize
    \caption{EPIC Kitchens Seen Test S1 results. (\%)}
    \begin{tabular}{c c c c c c c c c c c c c c }
        \hline 
        \multirow{2}{*}{Team} & \multirow{2}{*}{Position} & \multicolumn{3}{c}{Top@1} & \multicolumn{3}{c}{Top@5} & \multicolumn{3}{c}{Precision} & \multicolumn{3}{c}{Recall}\tabularnewline
         &  & Verb & Noun & Action & Verb & Noun & Action & Verb & Noun & Action & Verb & Noun & Action\tabularnewline
        \hline 
        UTSBaidu & 1 & 70.41 & 52.85 & 42.57 & 90.78 & 76.62 & 63.55 & 60.44 & 47.11 & 24.94 & 45.82 & 50.02 & 26.93\tabularnewline
        NUS CVML & 2 & 66.56 & 49.60 & 41.59 & 90.10 & 77.03 & 64.11 & 59.43 & 45.62 & 25.37 & 41.65 & 46.25 & 26.98\tabularnewline
        FBK HuPBA & 3 & 68.68 & 49.35 & 40.00 & 90.97 & 72.45 & 62.23 & 60.63 & 45.45 & 21.82 & 47.19 & 45.84 & 24.34\tabularnewline
        \hline 
        \multicolumn{14}{c}{{[}...{]}}\tabularnewline
        \hline 
        EPIC & 37 & 39.00 & 13.93 & 6.01 & 77.03 & 33.98 & 16.90 & 16.33 & 5.59 & 1.10 & 12.19 & 6.47 & 1.32\tabularnewline
        SU & 38 & 41.98 & 8.50 & 4.44 & 75.31 & 21.29 & 10.84 & 13.95 & 7.14 & 1.61 & 13.22 & 5.40 & 1.12\tabularnewline
        VPULab & 39 & 24.12 & 6.06 & 2.55 & 58.47 & 16.35 & 6.71 & 12.67 & 8.73 & 2.70 & 8.49 & 6.11 & 1.41\tabularnewline
        \hline 
    \end{tabular}
    \label{tab:Seen Kitchens Results}
    \par
    \end{centering}
\end{table*}

\begin{table*}[!t]
    \begin{centering}
    \renewcommand{\arraystretch}{1.2}
    \footnotesize
    \caption{EPIC Kitchens Unseen Test S2 results. (\%)}
    \begin{tabular}{c c c c c c c c c c c c c c }
        \hline 
        \multirow{2}{*}{Team} & \multirow{2}{*}{Position} & \multicolumn{3}{c}{Top@1} & \multicolumn{3}{c}{Top@5} & \multicolumn{3}{c}{Precision} & \multicolumn{3}{c}{Recall}\tabularnewline
         &  & Verb & Noun & Action & Verb & Noun & Action & Verb & Noun & Action & Verb & Noun & Action\tabularnewline
        \hline 
        UTSBaidu & 1 & 60.43 & 37.28 & 27.96 & 83.07 & 63.67 & 46.81 & 35.23 & 32.60 & 17.35 & 28.97 & 32.78 & 19.82\tabularnewline
        GT WISC MPI & 2 & 60.05 & 38.14 & 27.35 & 81.97 & 63.81 & 45.24 & 33.59 & 31.94 & 16.52 & 29.30 & 33.91 & 20.05\tabularnewline
        NUS CVML & 3 & 54.56 & 33.46 & 26.97 & 80.40 & 60.98 & 46.43 & 33.60 & 30.54 & 14.99 & 25.28 & 28.39 & 17.97\tabularnewline
        \hline 
        \multicolumn{14}{c}{{[}...{]}}\tabularnewline
        \hline 
        EPIC & 37 & 37.28 & 11.85 & 4.75 & 71.56 & 28.41 & 14.82 & 14.93 & 2.93 & 1.17 & 11.60 & 6.26 & 2.05\tabularnewline
        SU & 38 & 33.05 & 4.88 & 2.15 & 66.17 & 14.20 & 6.21 & 9.01 & 2.33 & 0.99 & 9.23 & 3.30 & 0.99\tabularnewline
        VPULab & 39 & 18.09 & 3.28 & 1.02 & 49.61 & 13.11 & 3.45 & 6.74 & 3.11 & 0.96 & 5.00 & 3.28 & 0.64\tabularnewline
        \hline 
    \end{tabular}
    \label{tab:Unseen Kitchens Results}
    \par
    \end{centering}
\end{table*}

Inspired by the approaches in \cite{wang2019baidu, wu2019long}, we re-weight action probabilities \(p_a\) by using training priors. These priors represent the probability \(p_p(v,n)\) of a pair verb-noun being an action in the training set. Unobserved actions as \textit{"peeling knife"} have \(p_p(v,n) = 0\) whereas observed actions as \textit{"open door"} have \(p_p(v,n) = 1\). Final action probabilities are computed as:

\begin{equation}
    p_a = p_p(v,n) \odot (p_v \otimes p_n),
\end{equation}

where \((\odot)\) represents the Hadamard product.

\subsection{Training Procedure}
Given that the Challenge needs separate predictions for verb, noun and action, we have decided to train the action classification CNN in a multi-task fashion. The final loss guiding the training process is computed as:

\begin{equation} \label{eq:loss}
    L= L_a + L_v + L_n,
\end{equation}

with \(L_a\), \(L_v\) and \(L_n\) being action loss, verb loss and noun loss respectively. The three losses are computed using regular Negative Log-Likelihood (NLL) loss using \(p_a\), \(p_v\) and \(p_n\). The influence of the \(L_a\) loss is explored in the Results Section.

\section{Experiments}
\subsection{Implementation} \label{subsec:Implementations}
\subsubsection{Training} \label{subsec:Training}
For the spatial domain each input image is adapted to the network input by resizing the smaller edge to 256 and then randomly cropping to a square shape of \(224 \times 224\). Data augmentation via regular horizontal flips is also used.

For the temporal domain, the number of unsupervised \textit{chunks} has been fixed to \(4\). From each of them, \(6\) consecutive frames are randomly extracted resulting in a total of \(24\) non-uniformly sampled frames along the video sequence.

In this submission, due to computational limitations, we have just performed a prospective study, using for training only a subset of \(3600\) from the \(28.561\) training sequences (i.e. a \(12.60 \%\))\footnote{The specific training subset used for the study is available at: \href{https://github.com/vpulab/EPIC-Kitchens-2020-Subset}{EPIC Kitchens 2020 Subset}}. Final results on the whole set of test sequences defined in the challenge are highly biased by this issue. 

To minimize the loss function in Equation \ref{eq:loss} and optimize the network's trainable parameters, the regular Stochastic-Gradient-Descend with Momentum (SGD) algorithm is used. In all our experiments the initial learning rate was set to \(0.1\), Momentum was set to \(0.9\) and weight decay was set to \(0.0001\). Learning rate was decayed every 200 epochs by \(1e-1\). Finally, batch size was set to \(32\) video sequences.

\subsubsection{Inference}
Following common practice, given a test sequence and its \(4\) \textit{chunks}, we uniformly sample \(5\) clips for each chunk along its temporal axis. For each clip, we scale the shorter spatial side to 256 pixels and take
\(5\) crops of \(224 \times 224\) to cover the spatial dimension. This results in \(25\) different views per chunk and so, \(100\) views per test sequence. We average the softmax scores for the final prediction.

We report action, verb and noun prediction probabilities for the challenge as explained in Section \ref{subsec:CNN Architecture}.

\subsection{Main Results}
As stated in Section \ref{subsec:Training}, due to computational limitations we use a subset of the training set for training and validation purposes. Specifically, from the \(28.561\) available sequences, we selected \(4500\). An \(80 \%\) of these sequences were dedicated for training while the rest was used for validation. These division leads to \(3600\) training sequences which represents a \(12.60 \%\) of the available training data. Given this fact, the following prospective study is a preliminary set of experimental results which may validate our starting hypothesis. Future work will extend and complete this work by using all the available training data.

\subsubsection{Preliminary Experiments}
The aim of this section is to gauge the influence of the ego-motion compensation approach and the designed shared CNN architecture. To this aim, we have uses the original Temporal branch architecture from SlowFast \cite{feichtenhofer2019slowfast} as a baseline. This CNN is fed by non-compensated temporal windows of \(24\) consecutive frames from EPIC Kitchens sequences. To present a fair comparison with the proposed approach, training and inference details from Section \ref{subsec:Implementations} are applied in the same manner to the baseline. Results are presented in Table \ref{tab:Preliminary Results}.

Comparing results obtained by the baseline (non compensated videos, fixed temporal window) with the proposed \textit{Ours} method (compensated videos, non-uniform temporal sampling) we observe an increase in performance. This preliminary experiment suggests that the use of compensated sequences along with the non-overlapping sampling, while implemented via a highly improvable approach, increases Top@1 results by a \(1.25 \%\), a \(2.12 \%\) and a \(1.89 \%\) for action, verb and noun respectively. 

\subsubsection{2020 Challenge Results}
Although results are highly biased due to the reduced amount of used training data, we report 2020 Challenge results for both Seen (S1) and Unseen (S2) tests sets in Tables \ref{tab:Seen Kitchens Results} and Table \ref{tab:Unseen Kitchens Results} respectively.

\section{Conclusions}
This participation describes a novel approach for action recognition in egocentric videos based on camera motion compensation and a non-uniform temporal sampling. To this aim, ego-motion is estimated for a given video sequence. To overcome the problems of motion compensation in sequences with high camera motion, each video sequence is partitioned into temporal \textit{chunks} using unsupervised clustering. Camera motion is compensated for each chunk independently leading to subsequences with a quasi-static background. Compensated \textit{chunks} are then fed to a shared CNN to obtain features from each of them. Late fusion gathers features from all the chunk expanding the temporal receptive field of the CNN.

We have performed a prospective study on a limited set of training data from the whole EPIC Kitchens Dataset. Preliminary results indicate that the proposed approach, while implemented in a highly improvable way, increases performance with respect to a baseline based on non-compensated sequences and a fixed temporal window of analysis.

Future work will continue exploring this line of research besides of using the whole training set available.

{\small
\bibliographystyle{ieee_fullname}
\bibliography{egbib}

\begin{thebibliography}{10}\itemsep=-1pt

\bibitem{carreira2018short}
Joao Carreira, Eric Noland, Andras Banki-Horvath, Chloe Hillier, and Andrew
  Zisserman.
\newblock A short note about kinetics-600.
\newblock {\em arXiv preprint arXiv:1808.01340}, 2018.

\bibitem{damen2018scaling}
Dima Damen, Hazel Doughty, Giovanni Maria~Farinella, Sanja Fidler, Antonino
  Furnari, Evangelos Kazakos, Davide Moltisanti, Jonathan Munro, Toby Perrett,
  Will Price, et~al.
\newblock Scaling egocentric vision: The epic-kitchens dataset.
\newblock In {\em Proceedings of the European Conference on Computer Vision
  (ECCV)}, pages 720--736, 2018.

\bibitem{dusmanu2019d2}
Mihai Dusmanu, Ignacio Rocco, Tomas Pajdla, Marc Pollefeys, Josef Sivic,
  Akihiko Torii, and Torsten Sattler.
\newblock D2-net: A trainable cnn for joint detection and description of local
  features.
\newblock In {\em Proceedings of the IEEE International Conference on Computer
  Vision (CVPR)}, 2019.

\bibitem{feichtenhofer2019slowfast}
Christoph Feichtenhofer, Haoqi Fan, Jitendra Malik, and Kaiming He.
\newblock Slowfast networks for video recognition.
\newblock In {\em Proceedings of the IEEE International Conference on Computer
  Vision (CVPR)}, pages 6202--6211, 2019.

\bibitem{fischler1981random}
Martin~A Fischler and Robert~C Bolles.
\newblock Random sample consensus: a paradigm for model fitting with
  applications to image analysis and automated cartography.
\newblock {\em Communications of the ACM}, 24(6):381--395, 1981.

\bibitem{hartley2003multiple}
Richard Hartley and Andrew Zisserman.
\newblock {\em Multiple view geometry in computer vision}.
\newblock Cambridge university press, 2003.

\bibitem{kanungo2002efficient}
Tapas Kanungo, David~M Mount, Nathan~S Netanyahu, Christine~D Piatko, Ruth
  Silverman, and Angela~Y Wu.
\newblock An efficient k-means clustering algorithm: Analysis and
  implementation.
\newblock {\em IEEE transactions on pattern analysis and machine intelligence},
  24(7):881--892, 2002.

\bibitem{kay2017kinetics}
Will Kay, Joao Carreira, Karen Simonyan, Brian Zhang, Chloe Hillier, Sudheendra
  Vijayanarasimhan, Fabio Viola, Tim Green, Trevor Back, Paul Natsev, et~al.
\newblock The kinetics human action video dataset.
\newblock {\em arXiv preprint arXiv:1705.06950}, 2017.

\bibitem{price2019evaluation}
Will Price and Dima Damen.
\newblock An evaluation of action recognition models on epic-kitchens.
\newblock {\em arXiv preprint arXiv:1908.00867}, 2019.

\bibitem{sigurdsson2016hollywood}
Gunnar~A Sigurdsson, G{\"u}l Varol, Xiaolong Wang, Ali Farhadi, Ivan Laptev,
  and Abhinav Gupta.
\newblock Hollywood in homes: Crowdsourcing data collection for activity
  understanding.
\newblock In {\em European Conference on Computer Vision}, pages 510--526.
  Springer, 2016.

\bibitem{wang2018non}
Xiaolong Wang, Ross Girshick, Abhinav Gupta, and Kaiming He.
\newblock Non-local neural networks.
\newblock In {\em Proceedings of the IEEE conference on computer vision and
  pattern recognition}, pages 7794--7803, 2018.

\bibitem{wang2019baidu}
Xiaohan Wang, Yu Wu, Linchao Zhu, and Yi Yang.
\newblock Baidu-uts submission to the epic-kitchens action recognition
  challenge 2019.
\newblock {\em arXiv preprint arXiv:1906.09383}, 2019.

\bibitem{wu2019long}
Chao-Yuan Wu, Christoph Feichtenhofer, Haoqi Fan, Kaiming He, Philipp
  Krahenbuhl, and Ross Girshick.
\newblock Long-term feature banks for detailed video understanding.
\newblock In {\em Proceedings of the IEEE Conference on Computer Vision and
  Pattern Recognition}, pages 284--293, 2019.

\end{thebibliography}
}

\end{document}